\documentclass[conference]{IEEEtran}
\IEEEoverridecommandlockouts

\usepackage{cite}
\usepackage{amsmath,amssymb,amsfonts}
\usepackage{algorithmic}
\usepackage{graphicx}
\usepackage{textcomp}
\usepackage{xcolor}
\def\BibTeX{{\rm B\kern-.05em{\sc i\kern-.025em b}\kern-.08em
    T\kern-.1667em\lower.7ex\hbox{E}\kern-.125emX}}
\begin{document}

\title{LAYERWISE TUNABLE LIFTING SCHEME FOR THE CONVOLUTIONAL NEURAL NETWORK\\
\thanks{This work was supported by the Institute of Information \& Communications Technology Planning \& Evaluation (IITP) – Innovative Human Resource Development for Local Intellectualization program grant funded by the Korea government (MSIT) (IITP-2026-RS-2020-II201741).}
}

\author{
\IEEEauthorblockN{
Abdumannon Yovkochov\textsuperscript{1},
An Le\textsuperscript{1},
Sungbal Seo\textsuperscript{2},
You-Suk Bae\textsuperscript{2},
Truong Nguyen\textsuperscript{1}
}
\IEEEauthorblockA{
\textsuperscript{1}Electrical and Computer Engineering Department, University of California San Diego, La Jolla, CA 92093, USA\\
\{ayovkochov, d0le, tqn001\}@ucsd.edu
}
\IEEEauthorblockA{
\textsuperscript{2}Department of Computer Engineering, Tech University of Korea, Siheung 15073, Korea\\
\{sungbal, ysbae\}@tukorea.ac.kr
}
}
\maketitle

\begin{abstract}
This work introduces a family of tunable lifting schemes for biorthogonal wavelet filter banks. We propose three lifting strategies: low-pass tuning (LS-LayLatt-LP), high-pass tuning (LS-LayLatt-HP), and a sequential lifting scheme that jointly adapts low- and high-frequency branches (LS-LayLatt-Sequential). All proposed designs are formulated using a lattice-based lifting structure\cite{biorUwU}, which guarantees invertibility and stability for arbitrary parameter values within the lifting functions.  We evaluated the proposed methods by integrating them into a ResNet-18\cite{resnet} backbone for image classification on the Describable Textures Dataset (DTD)\cite{dtdDataset}, as well as for anomaly detection on hazelnut images from the MVTec-AD\cite{mvtec} dataset and private KRC102S dataset. Experimental results demonstrate consistent performance improvements across all evaluated tasks.
\end{abstract}


\section{Introduction}
Convolutional Neural Networks (CNNs) such as ResNet~\cite{resnet}, VGG~\cite{vgg} and DenseNet~\cite{densenet} have played a major role in computer vision tasks. Downsampling operations, such as max pooling, average pooling, and strided convolution, are considered among the most important components of these architectures~\cite{shiftInvariantAgain}. However, these operations are deterministically low-pass, which leads to the loss of fine-grained information, introducing aliasing effects~\cite{shiftInvariantAgain}. To address these aliasing effects, many architectures have been proposed that use frequency-domain~\cite{frequnecyCNN, frequencyCNN2} or wavelet-based approaches~\cite{wavecnet1} as alternatives. Nevertheless, these methods typically utilize only low-frequency components or retain only low-pass information. As emphasized in Fig.~1, high-frequency components of images contain critical information for classification and detection. Consequently, Wavelet-Attention~\cite{wavelet-attention}, OrthLatt-UwU~\cite{orthlatt}, LS-BiorUwU\cite{biorUwU}, BiorLatt-UwU~\cite{BiorLatt} models were proposed to incorporate high-frequency information into the network. 
\begin{figure}[t]
    \centering
    \includegraphics[width=0.5\textwidth]{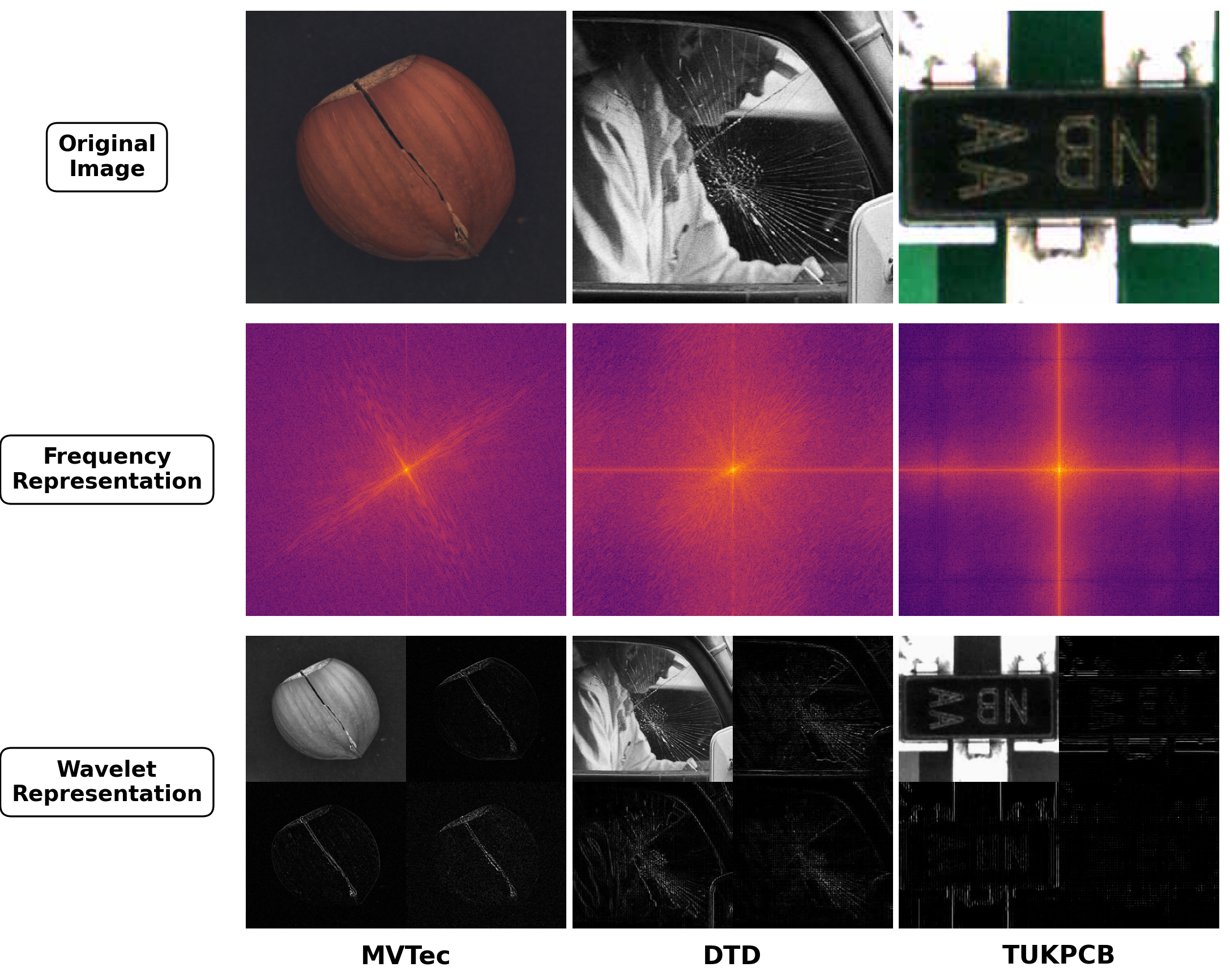}
    \caption{Wavelet (Haar) and frequency representations of samples from MVTecAD-hazelnut\cite{mvtec}, DTD-cracked\cite{dtdDataset}, and KRC102S (left to right). The figure highlights the role of high-frequency components in preserving discriminative visual information. In particular, for the DTD-cracked sample, retaining only the low-frequency component $\mathrm{X}_{ll}$  removes most crack structures, which are primarily encoded in high-frequency subbands. This illustrates how suppressing high-frequency details can lead to a significant loss of information critical for accurate classification and anomaly detection.}
    \label{fig:1}
\end{figure}

To address the aforementioned challenges and enhance flexibility, we propose a family of tunable biorthogonal wavelet lifting schemes for CNN downsampling operations. First, we investigate high-pass tuning to improve fine-detail preservation during the wavelet transform. In addition, low-pass tuning is studied to enhance coarse feature learning. Finally, we introduce a sequential lifting scheme that adaptively tunes both frequency branches while preserving perfect reconstruction. The proposed methods are integrated into a ResNet-18 backbone and evaluated on the DTD dataset~\cite{dtdDataset} for image classification, as well as on the MVTec-AD (hazelnut)~\cite{mvtec} and private KRC102S datasets for anomaly detection using CFLOW-AD~\cite{Cflow-ad}. These works have demonstrated that lattice-based parametrization and lifting schemes show an improvement on classification tasks compared to baseline methods. 

\section{Related Works}
Recent works have explored wavelet-based approaches for downsampling, pooling, and convolutional layers to improve performance on image classification and anomaly detection tasks. One of the earliest adopters of wavelet transforms for pooling layers is WaveCNet~\cite{wavecnet1}, which integrates discrete wavelet transforms (DWT) into CNN architectures using fixed orthogonal or biorthogonal wavelet families for DWT/IDWT. While effective, this fixed structure limits design flexibility. To address these limitations, subsequent works have proposed tunable and learnable wavelet constructions. For instance, OrthLatt-UwU~\cite{orthlatt} and BiorLatt-UwU~\cite{biorUwU} propose lattice-based orthogonal and biorthogonal wavelet units, respectively, with trainable coefficients and guaranteed perfect reconstruction, demonstrating notable improvements in computer vision tasks. However, orthogonal and biorthogonal wavelet structure used in OrthLatt-UwU\cite{orthlatt} and BiorLatt-UwU\cite{BiorLatt} restricts the filter design to equal filter lengths, limiting the ability to emphasize higher or lower frequency components. LS-BiorUwU\cite{biorUwU} attempts to tackle this problem by building wavelets using a lifting scheme which allows unequal lengths for high-pass and low-pass filters. Although the lifting scheme is used to tune the high-pass filter in the LS-BiorUwU\cite{biorUwU} method, the associated coefficients remain fixed by design throughout training.

In contrast, our method enables layer-wise tuning of lifting parameters, allowing each downsampling layer to learn its own wavelet parameters through cross-entropy–based optimization, while also supporting longer low-pass or high-pass filter lengths by design. This design allows different pooling layers to adapt their frequency responses according to their roles in the network hierarchy, leading to more effective feature extraction.
\section{Proposed Methods}
\begin{figure}[b]
    \centering
    \includegraphics[width=0.5\textwidth]{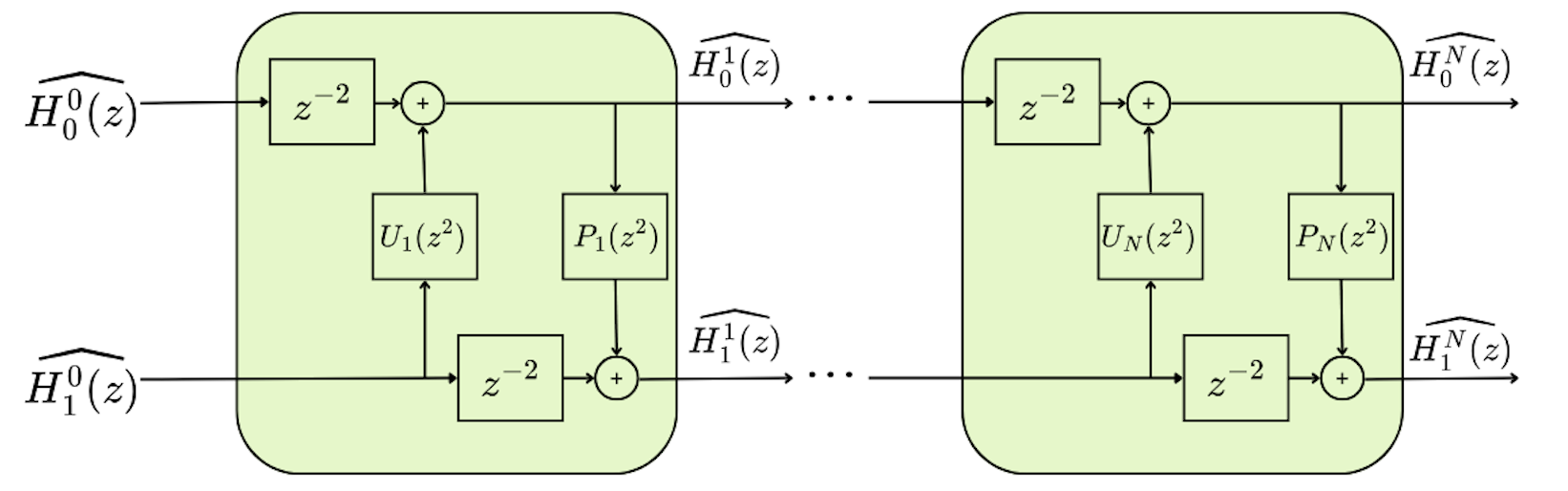}
    \caption{Lifting Scheme Structure for the analysis part of the filter bank, with lifting functions $U_k(z^2)$ and $P_k(z^2)$.} 
    \label{fig:2}
\end{figure}
\subsection{General Formulation}
Unlike standard orthogonal wavelets, in our work we use a biorthogonal wavelet, which relaxes orthogonality to biorthogonality, allowing filters to have unequal lengths. We use a lifting scheme to construct tunable biorthogonal wavelet filter banks. As illustrated in Fig. 2, the general form of lifting scheme consists of two lifting functions, $P_k(z)$ and $U_k(z)$, each responsible for tuning the high-pass and low-pass filters of the wavelet, respectively. We can construct the comprehensive lifting scheme for $K$ steps using the following equation:
\begin{equation}
\prod_{k=0}^{K-1}
\begin{bmatrix}
1 & 0 \\
P_k(z^2) & 1
\end{bmatrix}
\begin{bmatrix}
1 & U_k(z^2) \\
0 & 1
\end{bmatrix}
\end{equation}
where we define $P_k(z) = -a_k + a_k z^{-2k}$ and $U_k(z) = -b_k + b_k z^{-2k}$. Since both lifting matrices have unity on the diagonal, the determinant of each matrix is 1. Consequently, the determinant of the total lifting scheme is monomial, implying that the lifting structure is invertible. This ensures that, for any arbitrary choice of $a_k$ and $b_k$ in the lifting steps, the system satisfies the biorthogonality condition, and the synthesis filters can be exactly reconstructed by inverting the analysis operations in reverse order.

Additionally, as shown in Table~\ref{tab:filter_coefficients}, performing the lifting scheme does not break the finite impulse response (FIR) and linear-phase properties of the wavelet filters. Let $\widehat{H}_0^{k-1}(z)$ and $\widehat{H}_1^{k-1}(z)$ denote the analysis filters at lifting step $k-1$, and assume they are FIR and linear phase. Since the lifting scheme consists only of delays and finite-length filtering operations, finite support is preserved at every step. The newly constructed wavelet filter bank also preserves linear phase because each lifting operation modifies existing linear-phase filters through delayed linear combinations with real-valued coefficients, which introduce only constant phase shifts. The alignment delays in the lifting structure ensure a consistent phase center across all terms.

\begin{table}[t]
\caption{Filter coefficients after 1-step high-pass tuning}
\begin{center}
\begin{tabular}{|l|c|}
\hline
\textbf{Filter} & \textbf{Coefficients} \\ \hline
$h_0$ & $0.7071,\; 0.7071$ \\ \hline
$h_1$ & $-0.1183,\; -0.1183,\; 0.7071,\; -0.7071,\; 0.1183,\; 0.1183$ \\ \hline
\end{tabular}
\label{tab:filter_coefficients}
\end{center}
\end{table}

\subsection{High-Pass Tuning (LS-LayLatt-HP)}
We first explored a tunable lifting scheme which increases the high-pass filter length by using the low-pass filter. This method helps the high-pass filter achieve greater adaptability to high-frequency detail features of an image, such as texture and edges. By using the lifting structure mentioned above, we can construct this method by setting the parameters for $U_k = 0$, and assigning distinct learnable coefficients $a_k$ to the lifting function $P_k(z)$ for $k$ steps. We can represent this using the following recursive equation:
\begin{multline}
\begin{bmatrix}
\hat{H}_0^k(z) \\
\hat{H}_1^k(z)
\end{bmatrix}
=
\begin{bmatrix}
1 & 0 \\
P_k(z^2) & 1
\end{bmatrix}
\begin{bmatrix}
1 & 0 \\
0 & z^{-2}
\end{bmatrix}
\begin{bmatrix}
\hat{H}_0^{k-1}(z) \\
\hat{H}_1^{k-1}(z)
\end{bmatrix} \\
=
\begin{bmatrix}
\hat{H}_0^{k-1}(z) \\
- a_k \hat{H}_0^{k-1}(z)
+ z^{-2}\hat{H}_1^{k-1}(z)
+ a_k z^{-4k}\hat{H}_0^{k-1}(z)
\end{bmatrix}
\end{multline}
where $\widehat{H_0}$ and $\widehat{H_1}$ are the low-pass and high-pass filters of the wavelet filter bank, and $k$ is in the range from 1 to $N$.
\subsection{Low-Pass Tuning (LS-LayLatt-LP)}
We next investigate a complementary "lifting-down" scheme with tunable parameters, aimed at enhancing low-frequency representation learning within CNN architectures. While high-pass tuning emphasizes fine-scale and edge-related features, adaptive low-pass wavelet tuning is important for capturing coarse structures, global context, and long-range dependencies that are essential for robust feature hierarchies. Motivated by prior observations that tunable lifting parameters improve overall representation capacity, we extend the same principle to the low-pass branch of the wavelet filter bank.
We construct the lifting scheme for low-pass tuning in a manner similar to high-pass tuning. First, we set the coefficients for $P_k(z) = 0$ and assign a learnable parameter $b_k$ to the lifting function $U_k(z)$ for $k$ steps. Furthermore, we can express the entire lifting function using the following recursive form:
\begin{multline}
\begin{bmatrix}
\widehat{H}_0^{k}(z) \\
\widehat{H}_1^{k}(z)
\end{bmatrix}
=
\begin{bmatrix}
1 & U_k(z^2) \\
0 & 1
\end{bmatrix}
\begin{bmatrix}
z^{-2} & 0 \\
0 & 1
\end{bmatrix}
\begin{bmatrix}
\widehat{H}_0^{k-1}(z) \\
\widehat{H}_1^{k-1}(z)
\end{bmatrix} \\
=
\begin{bmatrix}
-b_k\widehat{H}_1^{k-1}(z) + z^{-2}\widehat{H}_0^{k-1}(z) + b_k z^{-4k}\widehat{H}_1^{k-1}(z) \\
\widehat{H}_1^{k-1}(z)
\end{bmatrix}
\end{multline}
\subsection{Sequential Tuning (LS-LayLatt-Sequential)}
Further, we explored a sequential lifting scheme in which the lifting operations are applied in a staged manner. Specifically, the low-pass branch of the wavelet transform is first updated using a tunable lifting step. The resulting updated low-pass representation is then used to guide a subsequent lifting operation that updates the high-pass branch. This sequential dependency introduces stronger coupling between low-frequency and high-frequency components, enabling more expressive and adaptive wavelet decompositions. This design helps construct longer and more flexible filters without explicitly increasing the filter size or violating the perfect reconstruction constraints of the wavelet filter bank. We construct the new wavelet filter bank using a recursive form below:
\begin{equation}
\begin{aligned}
\begin{bmatrix}\widehat{H}_0^{k}(z)\\ \widehat{H}_1^{k}(z)\end{bmatrix}
&=
\begin{bmatrix}1&0\\ P_k(z^2)&1\end{bmatrix}
\begin{bmatrix}1&0\\ 0&z^{-2}\end{bmatrix}
\begin{bmatrix}1&U_k(z^2)\\ 0&1\end{bmatrix}
\\
&\qquad\cdot
\begin{bmatrix}z^{-2}&0\\ 0&1\end{bmatrix}
\begin{bmatrix}\widehat{H}_0^{k-1}(z)\\ \widehat{H}_1^{k-1}(z)\end{bmatrix}
\\[2pt]
&=
\begin{bmatrix}
z^{-2}\widehat{H}_0^{k-1}(z) + U_k(z^2) \widehat{H}_1^{k-1}(z) \\
z^{-2}\widehat{H}_1^{k-1}(z) + P_k(z^2)\widehat{H}_0^{k}(z)
\end{bmatrix}
\end{aligned}
\end{equation}
for $k$ in the range from 1 to $N$, where $P_k(z) = -a_k + a_k z^{-2k}$ and $U_k(z) = -b_k + b_k z^{-2k}$, and where $a_k$ and $b_k$ are tunable coefficients for the lifting scheme.
\subsection{2D Implementation}
To integrate the proposed lifting schemes into CNN architectures, we extend the one-dimensional (1D) lifting formulation to the two-dimensional (2D) domain using a separable wavelet transform. The learned 1D low-pass and high-pass filters are applied successively along the horizontal and vertical dimensions of the input feature map. This process produces four subbands, $\mathbf{X}_{LL}$, $\mathbf{X}_{LH}$, $\mathbf{X}_{HL}$, and $\mathbf{X}_{HH}$, corresponding to the low--low, low--high, high--low, and high--high frequency components, respectively.
For a formulation suitable for deep learning, we construct low-pass $\mathbf{L}$ and high-pass $\mathbf{H}$ transform matrices from the learned filters $\widehat{H}_0^{\,k}$ and $\widehat{H}_1^{\,k}$ as
\begin{equation}
    \mathbf{L} = \mathbf{D}\mathbf{C}_{h_0}, \quad 
    \mathbf{H} = \mathbf{D}\mathbf{C}_{h_1}
    \label{eq:matrix_construction}
\end{equation}
where $\mathbf{D}$ denotes the downsampling matrix, and $\mathbf{C}_{h_0}$ and $\mathbf{C}_{h_1}$ are Toeplitz convolution matrices constructed from the coefficients of the low-pass and high-pass filters, respectively. Using these matrices, the 2D discrete wavelet transform can be expressed as
\begin{equation}
\begin{alignedat}{2}
\mathbf{X}_{LL} &= \mathbf{L}\mathbf{X}\mathbf{L}^{T}, \quad
&\mathbf{X}_{LH} &= \mathbf{L}\mathbf{X}\mathbf{H}^{T}, \\
\mathbf{X}_{HL} &= \mathbf{H}\mathbf{X}\mathbf{L}^{T}, \quad
&\mathbf{X}_{HH} &= \mathbf{H}\mathbf{X}\mathbf{H}^{T}
\end{alignedat}
\label{eq:2d_transform}
\end{equation}
where $\mathbf{X}$ denotes the input feature map.

\subsection{CNN Implementation}
The proposed methods are integrated into the ResNet family by replacing all conventional downsampling operations with the proposed tunable wavelet-based downsampling modules. Unlike fixed pooling layers in conventional CNNs, the lifting parameters in our method are trained jointly with the rest of the network using a cross-entropy loss function. This allows each downsampling layer to learn frequency responses that are optimized for the specific task for which it is used.
\section{Experiments and Results}
We integrated the proposed methods into the ResNet-18 architecture and evaluated different lifting schemes on the DTD dataset \cite{dtdDataset}. Furthermore, the resulting pretrained models were incorporated into the CFLOW-AD\cite{Cflow-ad} anomaly detection pipeline and applied to the MVTec-AD\cite{mvtec} hazelnut dataset as well as the KRC102S dataset. Across all datasets and experimental settings, the proposed models consistently outperformed the baseline methods.
\subsection{Image Classification: DTD}
To evaluate the texture classification capabilities of the proposed tunable lifting schemes, we use the DTD dataset\cite{dtdDataset}. The dataset consists of 5,640 high-resolution images across 47 texture categories. Given that texture features mainly consist of high-frequency components, DTD serves as a good benchmark to evaluate our methods' effectiveness. 

All experiments were conducted using a ResNet-18\cite{resnet} backbone. We compare our proposed LS-LayLatt (High-Pass, Low-Pass Tuning and Sequential Tuning) against multiple baseline models: 1) ResNet-18\cite{resnet} with the standard pooling layers, 2) WaveCNet\cite{wavecnet1} model with Resnet18 backbone, 3) LS-BiorUwU\cite{biorUwU} method which uses lifting scheme based method for high-pass filters, 4) the Orthogonal Lattice method (OrthLatt-UwU)\cite{orthlatt}, and 5) the Biorthogonal Lattice method with equal filter lengths (BiorLatt-UwU)\cite{BiorLatt}.
\begin{table}[t]
\caption{Classification accuracy (\%) on the DTD dataset over 5 random seeds for each measurement.}
\begin{center}
\begin{tabular}{|l|c|}
\hline
\multicolumn{2}{|c|}{\textbf{DTD}} \\ \hline
\textbf{Method} & \textbf{Accuracy (\%)} \\ \hline
ResNet-18 (Baseline)\cite{resnet} & 33.85 ($\pm 0.15$) \\ \hline
WaveCNet\cite{wavecnet1} & 26.70($\pm 0.17$) \\ \hline
LS-BiorUwU-3Steps\cite{biorUwU} & 43.50 ($\pm 0.21)$ \\ \hline
OrthLatt-UwU-4Taps\cite{orthlatt} & 44.51 ($\pm 0.48$) \\ \hline
BiorLatt-UwU-6Taps\cite{BiorLatt} & 45.63 ($\pm 0.37$)\\ \hline
LS-LayLatt-HP-1 Step (ours) & 42.59($\pm 0.25$) \\ \hline
LS-LayLatt-HP-2 Steps (ours) & 44.21 ($\pm 0.13$) \\ \hline
LS-LayLatt-HP-3 Steps (ours) & 45.32 ($\pm 0.37$) \\ \hline
LS-LayLatt-LP-1 Step (ours) & 42.79 ($\pm 0.18$)\\ \hline
LS-LayLatt-LP-2 Steps (ours) & 43.75 ($\pm 0.21$) \\ \hline
LS-LayLatt-LP-3 Steps (ours) & 43.42 ($\pm 0.41$) \\ \hline
LS-LayLatt-Sequential-1 Step (ours) & 45.27 ($\pm 0.28$) \\ \hline
LS-LayLatt-Sequential-2 Steps (ours) & \textbf{46.12 ($\pm$ 0.39)} \\ \hline
\end{tabular}
\label{tab:dtd_results_combined}
\end{center}
\end{table}
\subsubsection{LS-LayLatt Tuning Across different Strategies}
Table~\ref{tab:dtd_results_combined} summarizes the classification performance of different proposed lifting scheme strategies on the DTD dataset\cite{dtdDataset}.

We first focus on enhancing the high-frequency components of the   wavelet filter bank, which stores a significant
amount of information for texture classification. Compared to LS-BiorUwU-3Steps~\cite{biorUwU}   ($43.50\%$), our LS-LayLatt-HP introduces trainable lifting   coefficients and learns a distinct wavelet per layer, reaching   $45.32\%$, an $11.47$ point gain over ResNet-18 and an improvement over OrthLatt-UwU-4Taps~\cite{orthlatt}. Although the BiorLatt-UwU-6Taps~\cite{BiorLatt}  method achieves slightly higher accuracy, it relies on a globally shared wavelet across all pooling layers and also requires the equal length by design.

Furthermore, low-pass tuning (LS-LayLatt-LP) expands the low-frequency receptive   field and consistently outperforms ResNet-18~\cite{resnet}, but underperforms LS-LayLatt-HP, indicating   that high-frequency cues remain the dominant signal for texture   recognition. Sequential lifting (LS-LayLatt-Sequential) jointly extends   both branches and achieves the best overall result in   Table~\ref{tab:dtd_results_combined}, confirming that joint adaptation   of both frequency branches produces the most expressive filters.

\subsection{Anomaly Detection: MVTecAD and KRC102S}
For the anomaly detection task, we used the CFLOW-AD framework\cite{Cflow-ad}, a normalizing flow-based model, with modified or baseline Resnet18\cite{resnet} acting as a feature extractor. The feature extractor used for the CFLOW-AD framework\cite{Cflow-ad} was pretrained on the DTD\cite{dtdDataset} dataset to leverage texture-specific features. The model was evaluated on the public MVTec-AD Hazelnut\cite{mvtec} class and the private KRC102S dataset. 

We compare our proposed methods against five state-of-the-art baselines on the MVTec-AD Hazelnut~\cite{mvtec} dataset. The MVTec-AD Hazelnut dataset consists of images with various surface defects and contains 391 defect-free training samples and 110 test images (40 normal and 70 defective).

\begin{figure}[b]
    \centering
    \vspace{-3mm}
    \includegraphics[width=0.48\textwidth]{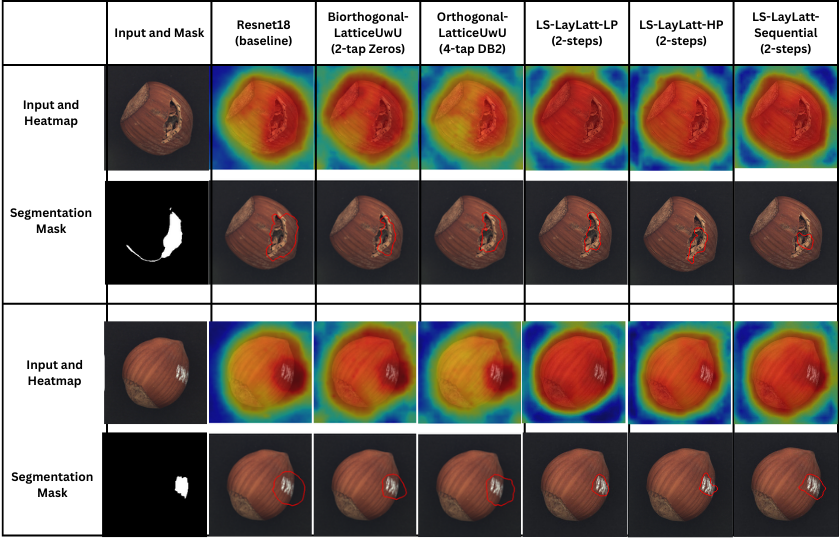}
    \vspace{-3mm}
    \caption{Segmentation and heatmaps generated by the CFLOW-AD method on the MVTec AD hazelnut dataset using a pretrained proposed and baseline methods with ResNet-18 backbone. }
    \label{fig:3}
\end{figure}

As demonstrated in Table~\ref{tab:mvtec_results}, our layer-specific tunable methods significantly outperform these baselines on MVTec-AD hazelnut. Specifically, the LS-LayLatt-Sequential (1-Step) method achieves the best image-level detection performance with an AUROC of 99.75\%, while the LS-LayLatt-LP (2-Steps) method outperforms all other approaches in the segmentation task. Qualitative results on the MVTec-AD Hazelnut dataset are shown in Fig. \ref{fig:3}, where anomaly heat maps and corresponding segmentation outputs are visualized. Although all compared methods are able to detect the anomalies, our proposed layer-specific tunable approaches produce highly localized segmentation maps. 
\begin{table}[t]
\caption{Detection (AUROC) and segmentation (AUROC) performance on the MVTEC-AD-hazelnut dataset over 5 random seeds for each measurement.}
\begin{center}
\begin{tabular}{|l|c|c|}
\hline
\multicolumn{3}{|c|}{\textbf{CFLOW-AD Mvtec-AD hazelnut}} \\ \hline
\textbf{Method} & \textbf{DET-AUROC} & \textbf{SEG-AUROC} \\ \hline
ResNet-18\cite{resnet} (Baseline) & 92.43 ($\pm$ 0.12) & 96.45 ($\pm$ 0.04) \\ \hline
WaveCNet\cite{wavecnet1} & 94.78 ($\pm$ 0.17) & 98.33 ($\pm$ 0.03) \\ \hline
LS-BiorUwU-3Steps\cite{biorUwU} & 95.47 ($\pm$ 0.12) & 97.56 ($\pm$ 0.02) \\ \hline
OrthLatt-UwU-4Taps\cite{orthlatt} & 90.11 ($\pm$ 0.21) & 97.17 ($\pm$ 0.05) \\ \hline
BiorLatt-UwU-6Taps\cite{BiorLatt} & 88.62 ($\pm$ 0.14) & 97.09 ($\pm$ 0.04) \\ \hline
LS-LayLatt-LP 2 Steps & 99.35 ($\pm$ 0.15) & \textbf{98.54 ($\pm$ 0.05)} \\ \hline
LS-LayLatt-HP 2 Steps & 98.37 ($\pm$ 0.11) & 98.31 ($\pm$ 0.02) \\ \hline
LS-LayLatt-Sequential 1 Step & \textbf{99.75 ($\pm$ 0.18)} & 98.47 ($\pm$ 0.07) \\ \hline
\end{tabular}
\label{tab:mvtec_results}
\end{center}
\end{table}

We further evaluated our methods on the KRC102S dataset, collected by Tech University of Korea. The dataset consists of Printed Circuit Board (PCB) component images and contains 4,346 anomaly-free training images and 248 test images (60 normal and 188 with various minor defects). Similar to the evaluation on the MVTec-AD Hazelnut dataset~\cite{mvtec}, we used a ResNet-18~\cite{resnet}-based encoder with different pooling methods as a feature extractor for the CFLOW-AD~\cite{Cflow-ad} model. Due to the absence of pixel-level ground-truth masks, evaluation was restricted to image-level detection AUROC. As shown in Table~\ref{tab:krc102s_results}, the low-pass tuning method underperformed compared to the ResNet-18 baseline; however, the sequential lifting scheme achieved the best performance, outperforming both the baseline and the other tunable configurations.

\begin{table}[t]
\caption{Anomaly Detection Accuracy for KRC102S over 3 random seeds for each measurement.}
\begin{center}
\begin{tabular}{|l|c|}
\hline
\multicolumn{2}{|c|}{\textbf{KRC102S-TUKPCB}} \\ \hline
\textbf{Method} & \textbf{DET-AUROC} \\ \hline
ResNet-18(Baseline) & 89.29 ($\pm 0.43$)\\ \hline
WaveCNet & 85.74 ($\pm 0.52$)\\ \hline
LS-LayLatt-LP 2 Steps & 88.81 ($\pm 0.47$)\\ \hline
LS-LayLatt-HP 2 Steps & 89.43 ($\pm 0.37$)\\ \hline
LS-LayLatt-Sequential 1 Step & \textbf{92.21} ($\pm \textbf{0.53}$)\\ \hline
\end{tabular}
\label{tab:krc102s_results}
\end{center}
\end{table}
\subsection{Lifting Parameters Analysis}
In this section, we examine the importance of parameter initialization for lifting-scheme coefficients. Since the lifting parameters are optimized using gradient-based methods with a cross-entropy loss, proper initialization is crucial for stable training and effective convergence. For all proposed methods, we initialize the wavelet filters using the Haar/Bior1.1\cite{strang1996} wavelet. For LS-LayLatt-HP, we use a coefficient initialization strategy similar to LS-BiorUwU~\cite{biorUwU}, where the lifting coefficients are initialized to approximate Bior1.3 and Bior1.5 wavelets for one and two lifting steps, respectively. This setup allows us to start from well-understood frequency response characteristics. For the lifting steps greater than two, additional coefficients are initialized with values close to zero, allowing the network to gradually learn higher-order refinements without introducing strong initial perturbations.

For LS-LayLatt-LP, the low-pass lifting coefficients are computed directly from the final filters obtained using LS-LayLatt-HP via the perfect-reconstruction relations between the analysis and synthesis filter banks. This approach provides a structured and effective initialization for texture-based classification tasks. In the sequential lifting scheme, the high-pass and low-pass branches are initialized using the coefficients learned from the LS-LayLatt-HP and LS-LayLatt-LP models, respectively. To avoid excessive filter growth, the number of lifting steps in the sequential scheme is limited to two.
\subsection{Computational Complexity and Experiment Setup}
The computational complexity of the learnable lifting scheme introduces only $2K$ learnable scalars per downsampling site for $K$ lifting steps and is therefore practically free in both parameters and floating-point operations. The observed overhead primarily from the $4{\times}$ channel widening required to fuse   the four wavelet subbands prior to the subsequent convolution, an architectural cost shared by all wavelet-pooling baselines~\cite{biorUwU,orthlatt,BiorLatt}.   Relative to the standard ResNet-18 backbone (11.7M parameters), the proposed LS-LayLatt variants reach 21.5M parameters, and these values are essentially independent of the   number of lifting steps $K$ since the per-site fusion convolution   dominates the cost. 

In total, we trained our LS-LayLatt models on the DTD dataset for 300 epochs, using a stage-wise transfer learning strategy \cite{transfer} with three stages of 100 epochs each. At the beginning of each stage, the model was initialized from the checkpoint achieving the best validation performance in the previous stage. Training was performed with a batch size of 4 using stochastic gradient descent. The initial learning rate was set to 0.01, and a step learning rate decay was applied every 30 epochs. Data augmentation was performed using random resized cropping and horizontal flipping during training. 

\section{Conclusion}
In conclusion, the integration of tunable lifting schemes into CNN architectures consistently demonstrates improved performance in both image classification and anomaly detection tasks. By incorporating learnable lifting parameters within biorthogonal wavelet filter banks, the proposed methods improve the network’s ability to capture both high-frequency and low-frequency features in a more task-specific manner. Each lifting-based method exhibits measurable performance gains compared to conventional architectures, confirming the effectiveness of wavelet-domain tuning for feature extraction. Moreover, the flexibility of the proposed lifting schemes, enables better spectral control without violating perfect reconstruction constraints. These results show that tuning wavelet lifting parameters provides an effective approach for improving representation learning and achieving better performance across different computer vision tasks. Additionally, future work will focus on extending the proposed lifting strategy to be able to support different base wavelet filter banks with varying filter lengths.

\bibliographystyle{IEEEtran}
\bibliography{refs}

@InProceedings{dtdDataset,
	      Author    = {M. Cimpoi and S. Maji and I. Kokkinos and S. Mohamed and and A. Vedaldi},
	      Title     = {Describing Textures in the Wild},
	      Booktitle = {Proceedings of the {IEEE} Conf. on Computer Vision and Pattern Recognition ({CVPR})},
	      Year      = {2014}
}

@INPROCEEDINGS{biorUwU,
  author={Le, An and Nguyen, Hung and Seo, Sungbal and Bae, You-Suk and Nguyen, Truong},
  booktitle={2025 33rd European Signal Processing Conference (EUSIPCO)}, 
  title={Biorthogonal Tunable Wavelet Unit with Lifting Scheme in Convolutional Neural Network}, 
  year={2025},
  volume={},
  number={},
  pages={1807-1811},
  doi={10.23919/EUSIPCO63237.2025.11226544}}

@ARTICLE{BiorLatt,
  author={Le, An D. and Jin, Shiwei and Seo, Sungbal and Bae, You-Suk and Nguyen, Truong Q.},
  journal={IEEE Open Journal of Signal Processing}, 
  title={Biorthogonal Lattice Tunable Wavelet Units and Their Implementation in Convolutional Neural Networks for Computer Vision Problems}, 
  year={2025},
  volume={6},
  number={},
  pages={768-783},
  doi={10.1109/OJSP.2025.3580967}}

@article{Cflow-ad,
  author       = {Denis A. Gudovskiy and
                  Shun Ishizaka and
                  Kazuki Kozuka},
  title        = {{CFLOW-AD:} Real-Time Unsupervised Anomaly Detection with Localization
                  via Conditional Normalizing Flows},
  journal      = {CoRR},
  volume       = {abs/2107.12571},
  year         = {2021},
  url          = {https://arxiv.org/abs/2107.12571},
  eprinttype    = {arXiv},
  eprint       = {2107.12571},
  bibsource    = {dblp computer science bibliography, https://dblp.org}
}

@InProceedings{vgg,
  author       = "Karen Simonyan and Andrew Zisserman",
  title        = "Very Deep Convolutional Networks for Large-Scale Image Recognition",
  booktitle    = "International Conference on Learning Representations",
  year         = "2015",
}

@INPROCEEDINGS{densenet,
  author={Huang, Gao and Liu, Zhuang and Van Der Maaten, Laurens and Weinberger, Kilian Q.},
  booktitle={2017 IEEE Conference on Computer Vision and Pattern Recognition (CVPR)}, 
  title={Densely Connected Convolutional Networks}, 
  year={2017},
  volume={},
  number={},
  pages={2261-2269},
  doi={10.1109/CVPR.2017.243}}

@inproceedings{mvtec,
  title={{MVTec AD}: A Comprehensive Real-World Dataset for Unsupervised Anomaly Detection},
  author={Bergmann, Paul and Fauser, Michael and Sattlegger, David and Steger, Carsten},
  booktitle={2019 IEEE/CVF Conference on Computer Vision and Pattern Recognition (CVPR)},
  pages={9584--9592},
  year={2019},
  doi={10.1109/CVPR.2019.00982},
  url={https://ieeexplore.ieee.org/document/8954181}
}

@ARTICLE{orthlatt,
  author={Le, An D. and Jin, Shiwei and Bae, You-Suk and Nguyen, Truong Q.},
  journal={IEEE Access}, 
  title={A Lattice-Structure-Based Trainable Orthogonal Wavelet Unit for Image Classification}, 
  year={2024},
  volume={12},
  number={},
  pages={88715-88727},
  doi={10.1109/ACCESS.2024.3418752}}

@ARTICLE{transfer,
  author={Pan, Sinno Jialin and Yang, Qiang},
  journal={IEEE Transactions on Knowledge and Data Engineering}, 
  title={A Survey on Transfer Learning}, 
  year={2010},
  volume={22},
  number={10},
  pages={1345-1359},
  doi={10.1109/TKDE.2009.191}}

@article{wavecnet1,
   title={WaveCNet: Wavelet Integrated CNNs to Suppress Aliasing Effect for Noise-Robust Image Classification},
   volume={30},
   ISSN={1941-0042},
   url={http://dx.doi.org/10.1109/TIP.2021.3101395},
   DOI={10.1109/tip.2021.3101395},
   journal={IEEE Transactions on Image Processing},
   publisher={Institute of Electrical and Electronics Engineers (IEEE)},
   author={Li, Qiufu and Shen, Linlin and Guo, Sheng and Lai, Zhihui},
   year={2021},
   pages={7074–7089} }

@book{strang1996,
  author    = {Gilbert Strang and Truong Nguyen},
  title     = {Wavelets and Filter Banks},
  publisher = {Wellesley--Cambridge Press},
  year      = {1996}
}

@inproceedings{shiftInvariantAgain,
  author    = {Richard Zhang},
  title     = {Making Convolutional Networks Shift-Invariant Again},
  booktitle = {Proceedings of the International Conference on Machine Learning (ICML)},
  year      = {2019}
}

@inproceedings{frequnecyCNN,
author = {Rippel, Oren and Snoek, Jasper and Adams, Ryan P.},
title = {Spectral representations for convolutional neural networks},
year = {2015},
publisher = {MIT Press},
address = {Cambridge, MA, USA},
booktitle = {Proceedings of the 29th International Conference on Neural Information Processing Systems - Volume 2},
pages = {2449–2457},
numpages = {9},
location = {Montreal, Canada},
series = {NIPS'15}
}

@article{frequencyCNN2,
  title={Learning Strides in Convolutional Neural Networks},
  author={Riad, Rachid and Teboul, Olivier and Grangier, David and Zeghidour, Neil},
  journal={ICLR},
  year={2022}
}

@misc{resnet,
      title={Deep Residual Learning for Image Recognition}, 
      author={Kaiming He and Xiangyu Zhang and Shaoqing Ren and Jian Sun},
      year={2015},
      eprint={1512.03385},
      archivePrefix={arXiv},
      primaryClass={cs.CV},
      url={https://arxiv.org/abs/1512.03385}, 
}

@misc{wavelet-attention,
      title={Wavelet-Attention CNN for Image Classification}, 
      author={Zhao Xiangyu},
      year={2022},
      eprint={2201.09271},
      archivePrefix={arXiv},
      primaryClass={cs.CV},
      url={https://arxiv.org/abs/2201.09271}, 
}
\end{document}